%% file: main.tex
\documentclass{article}

\usepackage[preprint]{neurips_2026}

\usepackage[utf8]{inputenc} 
\usepackage[T1]{fontenc}    
\usepackage{hyperref}       
\usepackage{url}            
\usepackage{booktabs}       
\usepackage{amsfonts}       
\usepackage{nicefrac}       
\usepackage{microtype}      

\usepackage{subcaption}
\usepackage{graphicx}
\usepackage{algorithm}
\usepackage{algpseudocode}
\usepackage{enumitem}
\usepackage{wrapfig}
\usepackage{floatflt}
\usepackage{amsthm}
\usepackage{xcolor}
\usepackage{todonotes}
\input{math_commands}

\title{DART: Distributional Adversarial Recurrent Training for Algorithm Learning}

\author{Hieu Tran Bao, Phung Thanh Dang \& Pham Quang Nhat Minh \\
FPT IS AI R\&D Center \\
Hanoi, Vietnam \\
\And
Hoang Thanh Tung \\
VNU -- University of Engineering and Technology \\
Hanoi, Vietnam
}

\begin{document}

\maketitle

\input{section/abstract}    
\input{section/introduction}
\input{section/related_work}
\input{section/preliminary}
\input{section/methodology}
\input{section/experiment}
\input{section/limitation}
\input{section/conclusion}

\newpage
\bibliography{neurips}
\bibliographystyle{unsrt}

\newpage
\appendix
\input{section/supplementary}
\input{section/ablation}

\end{document}

%% file: math_commands.tex
\usepackage{amsmath,amsfonts,bm}

\def\eqref#1{equation~\ref{#1}}

\def\1{\bm{1}}

\DeclareMathAlphabet{\mathsfit}{\encodingdefault}{\sfdefault}{m}{sl}
\SetMathAlphabet{\mathsfit}{bold}{\encodingdefault}{\sfdefault}{bx}{n}



%% file: section/abstract.tex
\begin{abstract}
Recurrent reasoning models (RRMs) can solve structured problems, achieving \textit{easy-to-hard} generalization through iterative computation in hidden space. 
These models are typically trained with instance-level supervision, which becomes increasingly problematic as task difficulty grows: valid solutions occupy a tiny region of the solution space, while invalid solutions proliferate rapidly.
We propose \textbf{Distributional Adversarial Recurrent Training (DART)}, a training framework that replaces single-point supervision with a local target distribution around the ground-truth solution and aligns model outputs with this distribution through an adversarial objective. 
DART provides a richer learning signal and encourages more stable iterative trajectories toward valid solutions.
When evaluated on Maze, Chess, and masked Sudoku with multiple RRMs, including Deep Thinking Systems and Tiny Recursive Models, DART improves solution quality, stability, and robustness under the evaluated distribution shifts. 
Comparisons with label smoothing, Gaussian softened targets, and progressive training show that DART is not explained by target softening alone and is complementary to training schemes that stabilize long-horizon recurrence.
These results identify DART as a promising approach for improving robustness across the evaluated recurrent reasoning models.
\end{abstract}

%% file: section/introduction.tex
\section{Introduction}

Logical reasoning has become a central benchmark for evaluating the reasoning capabilities of intelligent systems \cite{chollet2019measure, chollet2025arc0agi020}.
In large language model literature, this goal is often pursued through \textit{explicit reasoning} and \textit{inference-time search}.
LLM based approaches show that increasing inference-time computation---through sampling \cite{brown2024large}, branching \cite{wei2022chain0of0thought, yao2023tree}, and verification \cite{shao2025deepseekmath0v20}---can substantially improve performance on complex reasoning tasks.
These methods rely on large scale models to generate explicit reasoning traces and thus inherit the limitations of autoregressive models \cite{bachmann2025pitfallsnexttokenprediction}.

Parallel to these developments, another line of work explores \textit{implicit reasoning via iterative computation in hidden space}.
Recurrent reasoning models (RRMs), such as \textit{Tiny Recursive Models} (TRM) \cite{jolicoeur-martineau2025less} and \textit{Hierarchical Reasoning Models} (HRM) \cite{wang2025hierarchical}, can solve challenging logical reasoning tasks by repeatedly applying a fixed computation block, effectively scaling depth at test time. 
These results suggest that reasoning can emerge from \textit{iterative transformations of hidden states} without explicitly generating intermediate reasoning traces.
However, these models are primarily designed and trained on in-distribution datasets, and their performance on out-of-distribution data has not been systematically examined. 

Beyond logical reasoning and pattern matching, prior work \cite{DBLP:journals/corr/KaiserS15, graves2014neural} suggests that RRMs can learn \textit{algorithms} from data and exhibit systematic generalization capability.
Extending this idea, Schwarzschild et al. \cite{NEURIPS2021_3501672e} introduced the \textit{Deep Thinking System} (DTS), an RRM trained on relatively \textit{simple} problem instances that can generalize to substantially \textit{harder} ones at test time by increasing the number of recurrent reasoning steps.\footnote{
    A problem instance is considered harder than another instance if its shortest solution is longer than that of the other.
    We can create harder problems by increasing the problem sizes or decreasing the amount of information in problem statements.
    For example, a larger maze requires a longer shortest path to escape than a smaller maze, a 50\% masked Sudoku requires more steps to solve than a 30\% masked Sudoku.}
This phenomenon, referred to as \textit{easy-to-hard generalization}, suggests that models capable of genuine reasoning do not merely interpolate within the training distribution, but instead learn algorithmic procedures that can be systematically applied to more complex inputs.

Despite these promising results, RRMs are still typically optimized at the instance level: for each logical input, the model is trained to match a single target solution.
This creates a challenge for \textit{easy-to-hard} generalization because, as problem complexity increases, the number of incorrect solutions grows explosively while the set of valid solutions remains highly constrained.
As a result, the probability mass of correct solutions becomes vanishingly small relative to the vast space of invalid ones.
Consequently, iterative reasoning models often exhibit unstable performance on difficult inputs, requiring substantially more iterations and still failing to converge to correct solutions.

To mitigate this limitation of instance-level training, we introduce \textbf{Distributional Adversarial Recurrent Training (DART)} for Reasoning Models.
Inspired by latent-variable methods such as VAEs \cite{kingma2022autoencodingvariationalbayes}, our approach replaces single-point supervision with a target distribution that defines a neighborhood around the correct solution by injecting random noise during training.
This local soft target preserves the original solution under $\arg\max$ decoding, but it should not be interpreted as enumerating multiple distinct discrete solutions. Instead, it enlarges the supervised neighborhood around the observed solution in the probability simplex, making the learning signal less brittle than a single one-hot point.

To align generated reasoning states with this local target distribution, we introduce an \textbf{adversarial training objective} inspired by Wasserstein GANs (WGANs) \cite{arjovsky2017wgan}.
The reasoning model acts as a generator that produces candidate solution distributions, while a critic network learns to distinguish generated reasoning states from samples drawn from the soft target distribution.
By combining supervised reasoning loss with adversarial distribution alignment, our method encourages the model not only to produce correct solutions but also to remain close to an input-conditioned neighborhood of the target solution, thereby improving solution quality and trajectory stability on complex instances.

DART is generally applicable to different RRM architectures.
We evaluate it on both the Deep Thinking System (DTS) and the Tiny Recursive Model (TRM). 
For DTS, 
we consider two complementary easy-to-hard reasoning tasks: two-dimensional maze solving and chess puzzles. 
Across these tasks, DART improves DTS by increasing accuracy and yielding more stable performance across reasoning iterations.
Following \cite{jolicoeur-martineau2025less, wang2025hierarchical}, we evaluate DART on TRM for Sudoku. 
Because TRM is originally trained on hard Sudoku instances with a fixed masking level in both training and evaluation, it is not naturally formulated as an easy-to-hard benchmark. To study easy-to-hard generalization, we vary the Sudoku masking ratio, train the model on puzzles with low mask ratios, and evaluate on more heavily masked ones. 
Under this setting, DART also improves TRM's reasoning capability on more difficult Sudoku instances.

Overall, we summarize our contribution as follows:
\begin{itemize}
    \item To the best of our knowledge, we are the first to analyze \textit{easy-to-hard} generalization through the lens of solution-space imbalance. We argue that instance-level training exacerbates this difficulty as the imbalance becomes more severe as the complexity increases. 
    \item We reformulate the solution space of reasoning problems from a distributional perspective and propose \textbf{distributional adversarial recurrent training (DART)} to help recurrent reasoning models discover valid solutions more effectively.
    \item We show that DART improves the performance of multiple RRMs across a range of logical reasoning tasks, yielding more stable accuracy across reasoning iterations. We further compare against label smoothing, Gaussian softened targets without a critic, and progressive training baselines to separate target softening, adversarial alignment, and long-horizon stabilization.
\end{itemize}

%% file: section/related_work.tex
\section{Related Work}

\paragraph{Easy-to-hard generalization and algorithm learning.}
Sun et al. \cite{sun2024easy0to0hard} pose a fundamental question: if models are trained only on human-labeled data, how can they surpass human-level performance or solve problems that are harder than those seen during training? 
This challenge, termed \textit{easy-to-hard generalization}, has motivated research on enabling deep learning models to learn underlying algorithms for logical tasks and generalize to larger, more complex instances.
Recurrent neural networks (RNNs) and related architectures are well suited to this setting because they can flexibly adjust computational depth to input complexity \cite{article}. 
Graves et al. \cite{graves2014neural} propose the Neural Turing Machine (NTM), which mimics key behaviors of classical computers, enabling algorithm induction. 
The NTM can learn procedures such as sorting and copying and generalize to input sequences substantially longer than those seen during training. 
Neural GPU \cite{DBLP:journals/corr/KaiserS15} can learn arithmetic procedures such as addition and multiplication and generalize to arbitrary-length sequences.
Deep Thinking systems \cite{NEURIPS2021_3501672e} extend this direction by  iteratively refining intermediate representations through recurrent blocks. 
Original DTSs suffer from the \textit{overthinking} problem where more reasoning iterations lead to worse performance \cite{bansal2022end0to0end}.
To mitigate this issue, Bansal et al. \cite{bansal2022end0to0end} introduced both the recall architecture, which preserves input information across recurrent steps, and a progressive training objective that exposes the model to longer reasoning horizons during training.
Subsequent work further showed that progressive loss alone does not fully guarantee stable recurrence in compact DTS models; for small hidden representations, overthinking can persist unless the recurrent dynamics are more directly constrained, for example through Lipschitz-controlled architectures \cite{bear2024rethinking}.
We use the recall architecture as the backbone in all DTS experiments and explicitly compare against progressive training in Section~\ref{sec:experiment}.
More recent RRMs such as Tiny Recursive Models (TRM) \cite{jolicoeur-martineau2025less} and Hierarchical Reasoning Models (HRM) \cite{wang2025hierarchical} achieved strong results on challenging logical tasks while using far fewer parameters than LLMs. 
HRM uses a hierarchical architecture that combines slow and fast thinking to capture both local detail and global structure, whereas TRM simplifies this design with fewer parameters and emphasizes iterative refinement of model outputs.

\paragraph{Adaptive computation and iterative inference.}
Adjusting computational depth based on input complexity has been studied extensively in adaptive computation. 
Graves et al. (2016) propose Adaptive Computation Time (ACT) \cite{graves2016adaptive}, which enables recurrent models to halt once a learned stopping probability reaches a threshold. 
Universal Transformers \cite{dehghani2018universal} and Adaptive Recurrent Vision \cite{DBLP:conf/nips/VeerabadranRTRS23} applied this idea to a range of settings.
PonderNet \cite{banino2021pondernet0} improves the halting mechanism in ACT and outperforms ACT on logical extrapolation tasks such as prefix sums. 
More broadly, \cite{bansal2022end0to0end, kaya2018shallow0deep, Eyzaguirre_2020_CVPR} studied models with variable numbers of test-time iterations or layers. 
These methods are complementary to DART: they decide how much computation to allocate or how to train models over longer computation horizons, whereas DART changes the supervision signal used at each training example.

\paragraph{Adversarial guided training.}
Adversarial training has been widely adopted in generative modeling, beginning with Generative Adversarial Networks (GANs) \cite{goodfellow2014gan} and followed by advances such as WGAN \cite{arjovsky2017wgan}. 
Adversarial objectives have also been used to guide structured learning. 
SADM \cite{yang2024structureguidedadversarialtrainingdiffusion} allows diffusion models to capture semantic constraints among generated samples during denoising through adversarial training and batch-level similarity matrices. 
AdvBN \cite{shu2020encoding} improves robustness by perturbing feature statistics rather than input pixels. 
AdvBN generates worst-case perturbations of feature distributions (mean and variance) to simulate challenging style shifts and can be combined with existing data augmentation techniques to further improve performance.

\paragraph{Our contribution vs prior work.}
In this work, we identify a key challenge in easy-to-hard generalization as the severe imbalance between the valid and invalid solution spaces. 
By viewing logical reasoning tasks as a generation process, we argue that adversarial guided training can be applied to address this issue. 
Based on this perspective, we propose DART 
as a method for calibrating and regularizing the iterative reasoning process of RRMs.
DART does not modify the recurrent architecture or learn a halting policy. Instead, it replaces a single one-hot target with a local soft target distribution and uses an input-conditioned critic to align generated outputs with that local target neighborhood.

%% file: section/preliminary.tex
\section{Preliminary}


We consider logical reasoning tasks where the goal is to predict a structured solution $y$ from an input problem instance $x$. 
Examples include combinatorial puzzles, constraint satisfaction problems, and algorithmic reasoning tasks.

\textbf{Deep Thinking Systems (DTSs)} \cite{NEURIPS2021_3501672e} depart from classical recurrent architectures by producing the entire (representation of the) solution and refining that iteratively. 
DTSs process the input holistically, refine the solution through iterative computation, and learn algorithms end-to-end without supervision over individual intermediate steps. 
A DTS $f_\theta$ is defined as follows:
\begin{equation}
\phi_0 = p(x), \quad
\phi_{t+1} = r(\phi_t), \quad
\hat{y} = h(\phi_T)
\end{equation}
where $p$ is an input projection function, $r$ is a recurrent reasoning module applied for $T$ iterations, and $h$ is an output prediction head. 
A DTS can generalize to substantially harder/easier instances at test time by increasing/decreasing the number of reasoning iterations $T$ at test time. 
To prevent the model from forgetting the original problem during iterative computation, Bansal et al. \cite{bansal2022end0to0end} proposed the \textit{Recall} architecture, in which the original input $x$ is fed to every recurrent step:
\begin{equation}
\phi_{t+1} = r([\phi_t, x])
\end{equation}
In all DTS experiments in this paper, we use the Recall variant as the DTS backbone.

\textbf{Tiny Recursive Models (TRMs)} \cite{jolicoeur-martineau2025less} improve on the \textit{Hierarchical Reasoning Model} \cite{wang2025hierarchical}, which is built on the idea that structured reasoning requires iterative refinement and self-correction rather than a single large forward pass. 
This design is inspired by the human reasoning process, in which tentative hypotheses are revised multiple times before committing to an answer.
TRM constructs a first draft solution, repeatedly checks and refines it, and gradually moves toward a consistent final prediction. 
It maintains two latent states throughout inference: a \textit{low-level reasoning state} $z_t^{\ell}$, which serves as a flexible workspace for exploring constraints and local relationships, and a \textit{high-level answer state} $z_t^{h}$, which stores the current global hypothesis for the final solution. 
\begin{equation}
z_{t+1}^{\ell} = r_{\ell}(z_t^{\ell}, z_t^{h}, x), \quad
z_{t+1}^{h} = r_{h}(z_{t+1}^{\ell}, z_t^{h}), \quad
\hat{y} = h(z_T^{h}),
\end{equation}
where $r_{\ell}$ updates the low-level reasoning state, $r_h$ updates the high-level answer state, and $h$ maps the final high-level state to the output prediction. In practice, TRM reuses a single small computational block---typically a lightweight two-layer transformer or MLP---across many recursive steps, allowing the model to perform deep reasoning with a small number of parameters.

TRM alternates between multiple low-level refinement steps and a high-level update step. Intuitively, the model \textit{thinks} several times in the low-level workspace before revising its global answer hypothesis once, and this cycle is repeated until convergence. This separation between exploratory reasoning and answer consolidation enables TRM to solve structured tasks such as Sudoku through repeated draft-check-refine behavior rather than a single-pass prediction.

\textbf{Instance-level optimization.} Both DTS and TRM are trained to minimize a supervised loss between the predicted solution $\hat{y}$ at the final step $T$ and the ground-truth solution $y^*$:
\begin{equation}
\mathcal{L}_{sup} = \ell(\hat{y}, y^*)
\end{equation}
where $\ell$ is typically cross-entropy or mean squared error.
This optimization formulation treats the correct solution as a deterministic target and ignores the local structure of the valid solution space.

%% file: section/methodology.tex
\section{Proposed Method}
This section presents the details of our proposed \textbf{DART} framework. 
We show how to reformulate a reasoning problem as a distribution learning problem (Sec.~\ref{subsec:soft_solution}) and how to align the model's distribution with the target solution distribution (Sec.~\ref{subsec:adver_training}).

\subsection{Soft Solution Distribution}
\label{subsec:soft_solution}

We consider discrete logical reasoning problems with input $x$ and target solution $y^*$ such that $x, y^* \in \{c_0,c_1,\dots,c_k\}^{m \times n}$, 
where $c_0$ denotes a padding token and $c_1,\dots,c_k$ denote valid element values. 
The padding token allows us to handle problems where input and output have different sizes.
To transform a solution to a distribution, we first represent each solution element as a one-hot vector $v_{i,j} = \text{onehot}(y^*_{i,j}) \in \{0,1\}^{k+1}$.
The solution $y^*$ is represented as a 3D tensor
$v^* \in \{0,1\}^{m \times n \times (k+1)}$.
A reasoning model produces output $\hat{y} \in [0, 1]^{m \times n \times (k+1)}$ where each vector $y_{i,j} \in [0, 1]^{k+1}$ is a probability vector.
The discrete solution is obtained by applying the \textit{argmax} operator to $\hat{y}$.

At each training iteration, Gaussian noise is added to each one-hot vector to create a distribution around that vector:
\begin{equation}
\tilde{v}_{i,j} = v_{i,j}^* + \epsilon_{i,j}, \quad
\epsilon_{i,j} \sim \mathcal{N}(0, \sigma^2 I_{k+1}).
\end{equation}
To preserve the original solution under \textit{argmax} decoding, we restrict $\epsilon \in (-0.5, 0.5)$ and normalize the perturbed vector $\tilde{v}_{i,j}$ with softmax to produce vector $\hat{v}_{i, j}$ of the same form as $\hat{y}_{i,j}$.
After this operation, the target representation $\hat{v}$ and the model's output $\hat{y}$ lie in the same probability simplex, allowing generated predictions to be aligned directly with a local soft neighborhood around the observed solution. This construction preserves the original discrete target under $\arg\max$; it is not intended to represent all semantically distinct valid solutions when a problem admits multiple completions.


\subsection{Adversarial Solution Alignment}
\label{subsec:adver_training}

To align generated solutions with the local target distribution, we introduce an adversarial objective inspired by Wasserstein GANs.
Let $P_{\text{gen}}$ denote the distribution of solutions produced by the reasoning model $f_\theta$, and $P_{\text{real}}$ denote the soft solution distribution constructed around the ground-truth solution.

\subsubsection{Reasoning Critic Network}

A reasoning critic network $C_\psi$ with parameters $\psi$ assigns higher scores to softened target solutions and lower scores to generated ones. 
For each problem instance, the critic receives a concatenated input consisting of the original problem representation $x$ and a candidate soft solution $s$, where $s$ is either a generated solution $\hat{y}$ from the reasoning model or a softened target solution $\hat{v}$ sampled from the local target distribution. 
Its output is a scalar score that measures how well the candidate solution matches the input problem and conforms to the local target structure.
Formally, the critic input is defined as
$z = [x, s], s \in \{\hat{y}, \hat{v}\},$
where $[\cdot, \cdot]$ denotes concatenation along the feature dimension. 
To keep the overall framework simple and stable, we construct the critic with the same backbone architecture as the reasoning model. 
In other words, the critic reuses the same type of projection layer and recurrent reasoning block as the generator, and differs only in the final head, which outputs a scalar score rather than a structured solution tensor. 
Specifically, the critic is defined as follows:
\begin{equation}
\xi_0 = p_c(z), \quad
\xi_{t+1} = r_c(\xi_t), \quad
C_\psi(x, s) = q_c(\xi_T),
\end{equation}
where $p_c$ is an input projection layer, $r_c$ is a recurrent reasoning module, and $q_c$ maps the final hidden representation to a scalar score. 
Because the critic shares the same architectural form as the generator, it can evaluate candidate solutions using a compatible reasoning process while remaining lightweight to integrate into existing recurrent reasoning models. 
Intuitively, this allows the critic to assess whether a soft solution is globally consistent with the input instance while preserving architectural alignment between the two networks.
Our ablation study examines how the generator behaves when the critic is implemented as either a recurrent model or a feedforward model.

\subsubsection{Training Objective}
\paragraph{Critic optimization}
The critic is trained to maximize the Wasserstein distance between the two distributions:
\begin{equation}
\mathcal{L}_{critic}
=
-
\Big(
\mathbb{E}_{y \sim P_{\text{real}}}[C_\psi(x, \hat{v})]
-
\mathbb{E}_{\hat{y} \sim P_{\text{gen}}}[C_\psi(x, \hat{y})]
\Big)
\end{equation}
subject to a Lipschitz constraint on $C_\psi$. We enforce this constraint with critic weight clipping, following the original WGAN formulation. During this step, only the critic parameters $\psi$ are updated.

\paragraph{Reasoning model optimization}

The reasoning model is trained with both supervised learning and adversarial alignment. The adversarial loss encourages generated solutions to obtain higher critic scores:
\begin{equation}
\mathcal{L}_{adv}
=
-
\mathbb{E}_{\hat{y} \sim P_{\text{gen}}}[C_\psi(x, \hat{y})].
\end{equation}
This loss is optimized jointly with the supervised loss, resulting in the final training objective:

\begin{equation}
\mathcal{L}_{total}
=
\mathcal{L}_{sup}
+
\lambda \mathcal{L}_{adv}.
\end{equation}

During this step, the reasoning model parameters $\theta$ are updated while the critic parameters $\psi$ remain fixed.
The complete training procedure is summarized in Algorithm~\ref{alg:training}.

\begin{algorithm}[t]
\caption{Training Recursive Reasoning Models with Adversarial Alignment}
\label{alg:training}
\begin{algorithmic}[1]
\Require training data $(x, y^*)$, reasoning model $f_\theta$, critic $C_\psi$
\Require perturbation variance $\sigma$, adversarial weight $\lambda$
\While{not converged}

\State Sample minibatch $(x, y^*)$

\State // Construct soft solution distribution $P_{\text{real}}$
\State $v_{i,j} \leftarrow \text{onehot}(y^*_{i,j})$
\State $\tilde v_{i,j} \leftarrow v_{i,j} + \epsilon_{i,j}, \quad \epsilon_{i,j} \sim \mathcal{N}(0, \sigma^2 I)$
\State $\hat v_{i,j} \leftarrow \text{softmax}(\tilde v_{i,j})$
\State $y_{real} \leftarrow \{\hat v_{i,j}\}$
\State // Generate solution from recursive reasoning model
\State $\hat y \leftarrow f_\theta(x)$
\State // {Update critic}
\State Compute critic loss
\[
\mathcal{L}_{critic} =
-\Big(
\mathbb{E}_{y \sim P_{\text{real}}}[C_\psi(x, y)]
-
\mathbb{E}_{\hat y \sim P_{\text{gen}}}[C_\psi(x, \hat y)]
\Big)
\]
\State Update critic parameters
\[
\psi \leftarrow \psi - \eta \nabla_\psi \mathcal{L}_{critic}
\]
\State // {Update reasoning model}
\State Compute supervised loss
\[
\mathcal{L}_{sup} = \ell(\hat y, y^*)
\]
\State Compute adversarial loss
\[
\mathcal{L}_{adv} =
-
\mathbb{E}_{\hat y \sim P_{\text{gen}}}[C_\psi(x, \hat y)]
\]
\State Compute total loss
\[
\mathcal{L}_{total}
=
\mathcal{L}_{sup}
+
\lambda\mathcal{L}_{adv}
\]
\State Update reasoning model
\[
\theta \leftarrow \theta - \eta \nabla_\theta \mathcal{L}_{total}
\]
\EndWhile
\end{algorithmic}
\end{algorithm}

%% file: section/experiment.tex
\section{Experiments}
\label{sec:experiment}
\subsection{Datasets, model architecture and training}

Following Schwarzschild et al. \cite{NEURIPS2021_3501672e}, we evaluate DART on easy-to-hard reasoning benchmarks for Deep Thinking Systems (DTS). We additionally reformulate Sudoku \cite{wang2025hierarchical} as an easy-to-hard benchmark and use it to evaluate DART on Tiny Recursive Models (TRM) under partial observability. Detailed dataset descriptions and implementation details are provided in Appendix~\ref{appendix:datasets}.

\subsection{Results}
We organize the results around three questions. First, does DART improve over instance-level cross-entropy (CE) as the easy-to-hard shift becomes stronger, and does this hold beyond DTS? Second, how does DART interact with progressive training, which was designed to address long-horizon overthinking? Third, can the gains be explained by simpler target augmentations such as label smoothing or softened targets without adversarial alignment?

\begin{figure}[t]
    \centering
    \includegraphics[width=0.95\linewidth]{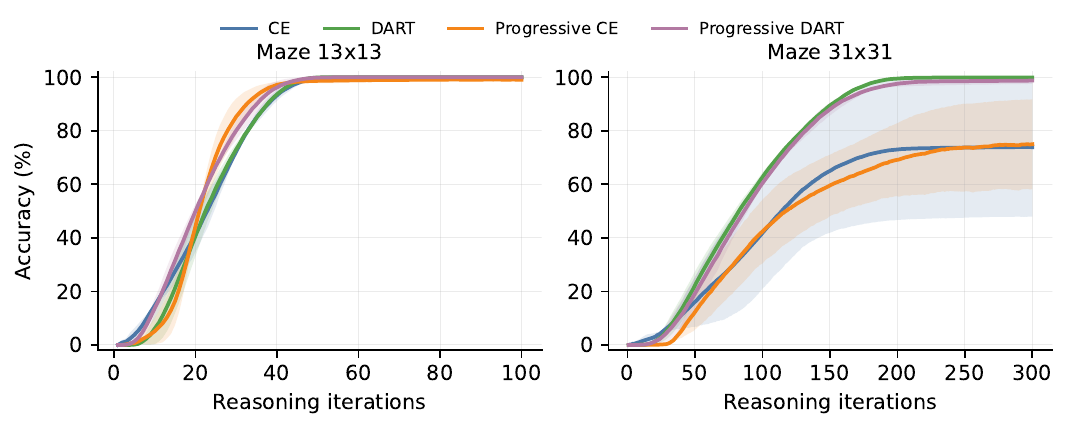}
    \caption{Maze accuracy as the evaluation size increases from $13\times13$ to $31\times31$. The CE failure mode is mostly absent at the smaller size but appears under the harder spatial extrapolation setting.}
    \label{fig:maze_scale_progressive}
\end{figure}

\begin{table}[t]
\centering
\small
\caption{Maze scaling results for standard and progressive training (mean $\pm$ std over seeds).}
\label{tab:progressive_maze}
\resizebox{\linewidth}{!}{%
\begin{tabular}{lcccc}
\toprule
Method & $13\times13$ Peak & $13\times13$ Final (100) & $31\times31$ Peak & $31\times31$ Final (300) \\
\midrule
CE & $99.88 \pm 0.00$ & $99.85 \pm 0.03$ & $73.92 \pm 26.01$ & $73.88 \pm 25.98$ \\
Progressive CE & $99.17 \pm 0.59$ & $99.06 \pm 0.63$ & $75.01 \pm 16.69$ & $74.96 \pm 16.75$ \\
DART & $\mathbf{100.00 \pm 0.00}$ & $\mathbf{100.00 \pm 0.00}$ & $\mathbf{99.91 \pm 0.13}$ & $\mathbf{99.91 \pm 0.13}$ \\
Progressive DART & $99.98 \pm 0.01$ & $99.98 \pm 0.01$ & $98.75 \pm 1.25$ & $98.73 \pm 1.26$ \\
\bottomrule
\end{tabular}
}
\end{table}

\begin{wrapfigure}{r}{0.5\textwidth}
    \centering
    \includegraphics[width=1.0\linewidth]{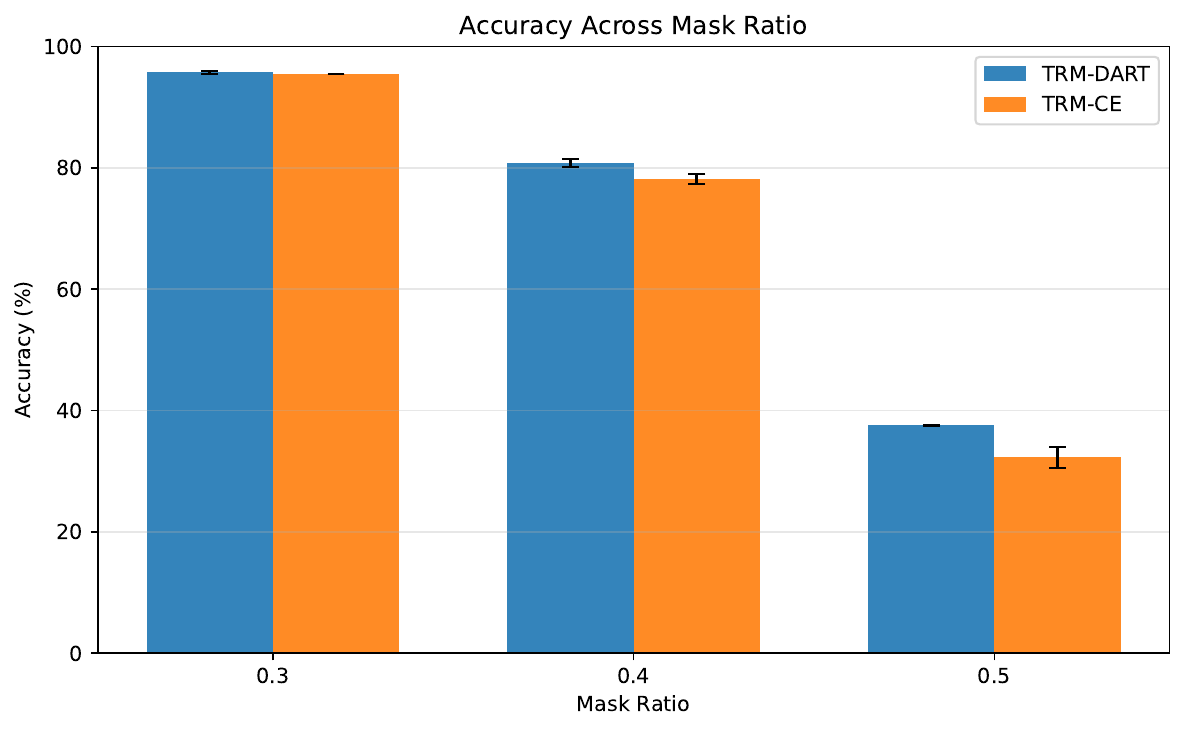}
    \caption{Sudoku accuracy under increasing masking ratios. DART consistently outperforms CE, with larger gains under stronger distribution shift.}
    \label{fig:sudoku_bar}
\end{wrapfigure}

\begin{table}[t]
\centering
\caption{Test accuracy (\%) on Sudoku-Extreme under different masking ratios.}
\label{tab:sudoku_mask}
\begin{tabular}{ccc|ccc|ccc}
\toprule
& & & \multicolumn{3}{c|}{\textbf{Standard CE}} & \multicolumn{3}{c}{\textbf{DART}} \\
\cmidrule(lr){4-6} \cmidrule(lr){7-9}
$\textit{H}$ & $\textit{L}$ & Depth & 30\% & 40\% & 50\% & 30\% & 40\% & 50\% \\
\midrule
3 & 6 & 42 & $95.5\pm0.0$ & $78.1\pm0.8$ & $32.3\pm1.7$ & $\mathbf{96.0\pm0.0}$ & $\mathbf{81.4\pm0.6}$ & $\mathbf{37.5\pm0.1}$ \\
4 & 6 & 56 & $95.5\pm0.0$ & $78.2\pm0.8$ & $32.4\pm1.7$ & $\mathbf{96.0\pm0.0}$ & $\mathbf{81.5\pm0.6}$ & $\mathbf{38.0\pm0.1}$ \\
5 & 6 & 70 & $95.5\pm0.0$ & $78.3\pm0.8$ & $32.5\pm1.7$ & $\mathbf{96.0\pm0.0}$ & $\mathbf{81.6\pm0.6}$ & $\mathbf{38.4\pm0.1}$ \\
6 & 6 & 84 & $95.5\pm0.0$ & $78.3\pm0.8$ & $32.6\pm1.7$ & $\mathbf{96.0\pm0.0}$ & $\mathbf{81.6\pm0.6}$ & $\mathbf{38.5\pm0.1}$ \\
\bottomrule
\end{tabular}
\end{table}

\subsubsection{DART improves over instance-level CE}
Maze exposes how the CE failure mode depends on the severity of the easy-to-hard shift. Models are trained on $9\times9$ mazes and evaluated at larger sizes. At $13\times13$, the shift is mild and CE already reaches $99.88 \pm 0.00\%$ peak accuracy and $99.85 \pm 0.03\%$ final accuracy (Table~\ref{tab:progressive_maze}). In this regime, the search space is only moderately larger than the training distribution, so point-wise supervision is still sufficient for the recurrent process to recover the correct path.

The same objective becomes unreliable at $31\times31$. CE reaches only $73.92 \pm 26.01\%$ peak accuracy, with a bimodal outcome across seeds: one run nearly solves the task, while another saturates around $47.9\%$. This indicates that the difficulty is not merely larger input size, but the rapid growth of locally plausible yet globally invalid path configurations. A one-hot target rewards only the exact final path and gives little guidance about nearby partial solutions or how to remain on the feasible path manifold over many recurrent updates. DART removes most of this sensitivity, reaching $99.91 \pm 0.13\%$ peak and final accuracy at $31\times31$. The local soft target provides a non-brittle neighborhood around the ground-truth path, while the input-conditioned critic penalizes generated outputs that drift away from maze structure.

Chess shows a different CE failure mode. The model is trained on the easiest 500K puzzles and evaluated on harder puzzles with indices from 600K to 700K. CE reaches a moderate peak of $80.81 \pm 0.37\%$ (Table~\ref{tab:progressive_chess}), but by iteration 50 collapses to $0.92 \pm 0.08\%$. Thus CE can briefly find useful move representations, but repeated recurrent refinement destroys them. DART improves peak solution quality to $84.83 \pm 0.03\%$, suggesting that adversarial alignment provides a more global scoring signal for a task where many moves are locally plausible but only a few are optimal. However, DART alone does not fully eliminate late-iteration overthinking on Chess, which motivates the progressive-training comparison below.

Sudoku tests whether the same principle transfers beyond DTS. We evaluate TRM \cite{jolicoeur-martineau2025less} by training with 20\% masking and testing on 10{,}000 puzzles with 30\%, 40\%, and 50\% masking. Table~\ref{tab:sudoku_mask} shows that DART consistently outperforms CE at every computation depth and masking ratio. The gap increases as the shift becomes harder: at depth 56, the gain is $0.5\%$ at 30\% masking, $3.3\%$ at 40\%, and $5.6\%$ at 50\%. Figure~\ref{fig:sudoku_bar} shows the same trend across masking ratios. Under heavier masking, the model must reason over a larger set of plausible completions rather than recover a nearly deterministic solution. DART improves robustness by aligning predictions toward a local neighborhood around the observed solution instead of a single hard target.

\begin{figure}[h]
    \centering
    \includegraphics[width=0.88\linewidth]{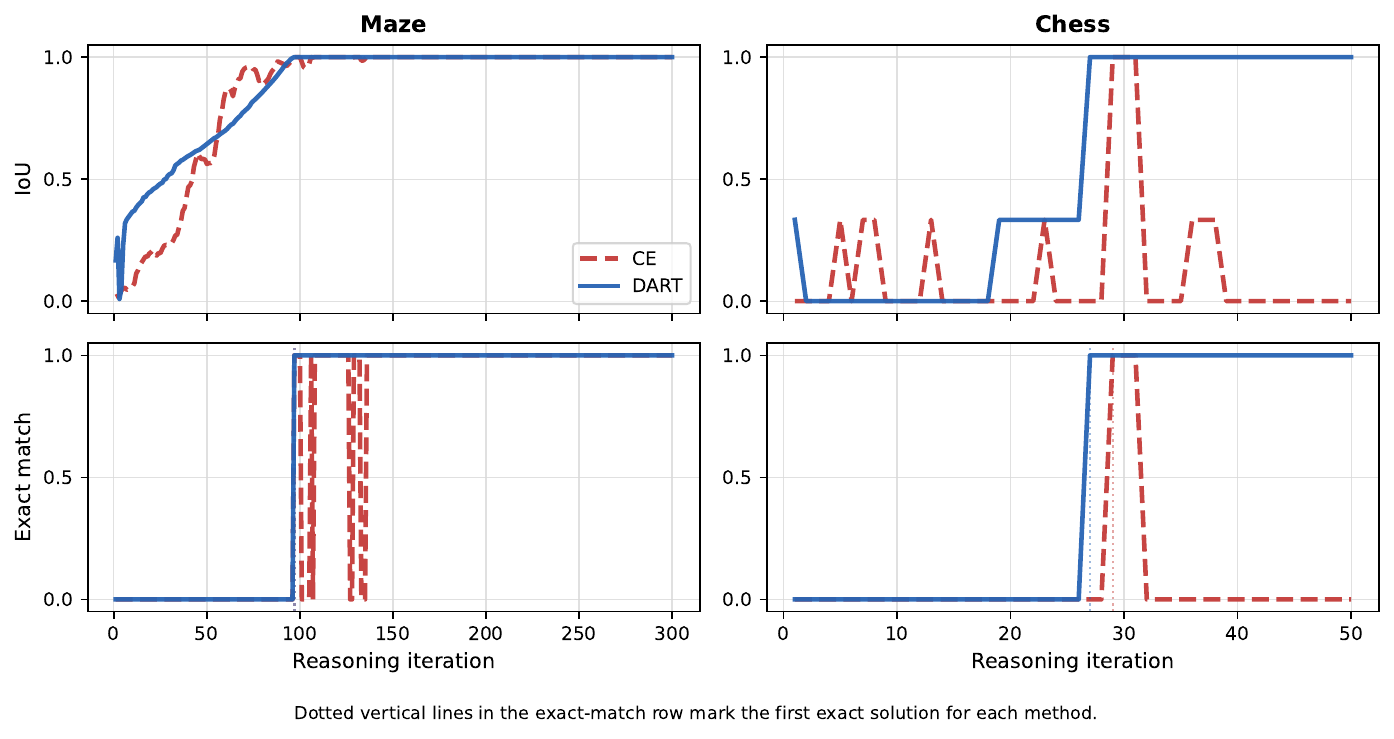}
    \caption{Sample reasoning trajectories for CE and DART on Maze and Chess. The top row shows IoU with the target solution, while the bottom row shows exact match; dotted vertical lines mark the first exact solution. In these examples, DART reaches the exact solution and remains there, whereas CE can enter the correct region transiently and then drift away.}
    \label{fig:sample_trajectory_metrics}
\end{figure}

\paragraph{Trajectory-level evidence.}
To inspect the mechanism behind these aggregate results, we compare representative CE and DART trajectories in Figure~\ref{fig:sample_trajectory_metrics}. IoU measures partial overlap with the target, while exact match identifies iterations where the full discrete solution is correct.

\begin{table}[h]
\centering
\small
\caption{Aggregate trajectory statistics over held-out samples (Maze: 128 samples; Chess: 512 samples). First exact iteration and retention are computed over samples that reach an exact solution at least once.}
\label{tab:trajectory_stats}
\resizebox{\linewidth}{!}{%
\begin{tabular}{llccccc}
\toprule
Task & Method & Solved once (\%) & Final exact (\%) & First exact iter & Retention after first exact (\%) & Exact-match AUC (\%) \\
\midrule
Maze $31\times31$ & CE & 51.6 & 43.8 & 88.6 & 82.0 & 28.4 \\
Maze $31\times31$ & DART & \textbf{100.0} & \textbf{100.0} & 94.0 & \textbf{99.9} & \textbf{68.9} \\
Chess & CE & 84.8 & 0.0 & 28.0 & 11.6 & 4.5 \\
Chess & DART & \textbf{91.0} & \textbf{15.8} & \textbf{23.1} & \textbf{63.5} & \textbf{32.3} \\
\bottomrule
\end{tabular}
}
\end{table}

On the representative Maze example, CE and DART both first reach exact match at iteration 97, but CE drops out of exact match 11 times afterward whereas DART remains exact for all subsequent iterations. This behavior holds across samples: DART solves and finishes all 128 evaluated $31\times31$ mazes, while CE solves only 51.6\% at least once and finishes at 43.8\% exact match (Table~\ref{tab:trajectory_stats}). On Chess, CE reaches exact match in many samples but almost never preserves it, ending with 0.0\% final exact match and only 4.5\% exact-match AUC. DART reaches exact solutions more often, reaches them earlier on average (23.1 vs. 28.0 iterations), and retains them much longer, increasing retention from 11.6\% to 63.5\% and exact-match AUC from 4.5\% to 32.3\%. This supports the hypothesis that the critic does not simply accelerate convergence, but helps keep recurrent outputs inside an input-conditioned neighborhood of successful traces.

\subsubsection{DART and progressive training are complementary}
Progressive training and DART address related but distinct failure modes. Progressive training was introduced for DTS to reduce \textit{overthinking}: it changes the training schedule so the same recurrent architecture experiences longer reasoning horizons and learns not to degrade when computation continues. DART instead changes the supervision signal by replacing a single one-hot target with a local soft target distribution and an input-conditioned adversarial constraint. Thus, progressive training primarily targets stability along the iteration axis for DTS, while DART targets imbalance in the valid solution distribution and can be applied at the objective level to different RRMs, as shown by our DTS and TRM experiments. This distinction matters for recursive architectures such as TRM and HRM, whose inference already contains nested low-level and high-level loops; directly porting the DTS progressive schedule to such models is not a natural drop-in baseline, whereas DART only requires a structured output distribution and a critic defined on the model's predictions.

This distinction also means that progressive training should not be expected to solve all easy-to-hard failures. Prior work on stable Deep Thinking models shows that progressive loss alone can still fail to prevent overthinking when the hidden representation is small, motivating architectural constraints on the recurrent map \cite{bear2024rethinking}. Our results show a complementary limitation: even when progressive training improves recurrent stability, it may not address the growth of invalid solution configurations under harder problem distributions.

\begin{figure}[h]
    \centering
    \includegraphics[width=0.62\linewidth]{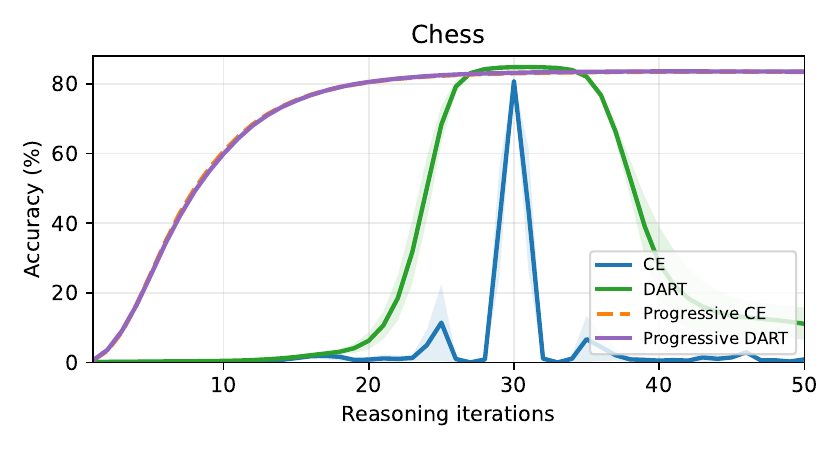}
    \caption{Progressive training baselines on Chess. Progressive training removes the late-iteration collapse of CE and DART, while DART alone reaches the highest peak accuracy.}
    \label{fig:chess_progressive}
\end{figure}

\begin{table}[h]
\centering
\small
\caption{Progressive training baselines on Chess (mean $\pm$ std over seeds).}
\label{tab:progressive_chess}
\resizebox{\linewidth}{!}{%
\begin{tabular}{lcccc}
\toprule
Method & Peak & Peak iter & Acc. @ iter 40 & Final (50) \\
\midrule
CE & $80.81 \pm 0.37$ & 30, 30 & $0.56 \pm 0.11$ & $0.92 \pm 0.08$ \\
Progressive CE & $83.50 \pm 0.03$ & 41, 45 & $83.48 \pm 0.02$ & $83.46 \pm 0.02$ \\
DART & $\mathbf{84.83 \pm 0.03}$ & 31, 31 & $28.58 \pm 10.47$ & $11.13 \pm 4.64$ \\
Progressive DART & $83.59 \pm 0.09$ & 41, 50 & $\mathbf{83.58 \pm 0.09}$ & $\mathbf{83.52 \pm 0.16}$ \\
\bottomrule
\end{tabular}
}
\end{table}

This distinction is clearest on Maze. At $13\times13$, all methods are near perfect because the CE failure mode has not yet appeared. At $31\times31$, Progressive CE reduces variance relative to CE but reaches only $75.01 \pm 16.69\%$ peak accuracy, far below DART's $99.91 \pm 0.13\%$ (Table~\ref{tab:progressive_maze}). This shows that making a DTS less erratic over longer horizons does not by itself solve the larger spatial extrapolation problem: the model still receives point-wise supervision in a solution space where invalid paths grow much faster than valid ones. Progressive DART remains strong at $98.75 \pm 1.25\%$, confirming that the two objectives are compatible, but DART alone already addresses the dominant Maze failure mode.

On Chess, the roles reverse. The main failure is late-iteration collapse: CE falls from an $80.81 \pm 0.37\%$ peak to $0.92 \pm 0.08\%$ at iteration 50, and DART, while reaching the best peak accuracy, still drops to $11.13 \pm 4.64\%$. Figure~\ref{fig:chess_progressive} and Table~\ref{tab:progressive_chess} show that Progressive CE largely removes this overthinking behavior and reaches $83.46 \pm 0.02\%$ at iteration 50. Progressive DART inherits the same long-horizon stability while retaining a distributional objective, achieving the best final accuracy, $83.52 \pm 0.16\%$. The results therefore support a complementary view: DART improves the quality of the solution distribution reached by the recurrent process, and progressive training keeps that quality from degrading as computation continues.

\begin{figure}[t]
    \centering
    \begin{subfigure}{0.45\textwidth}
        \centering
        \includegraphics[width=\textwidth]{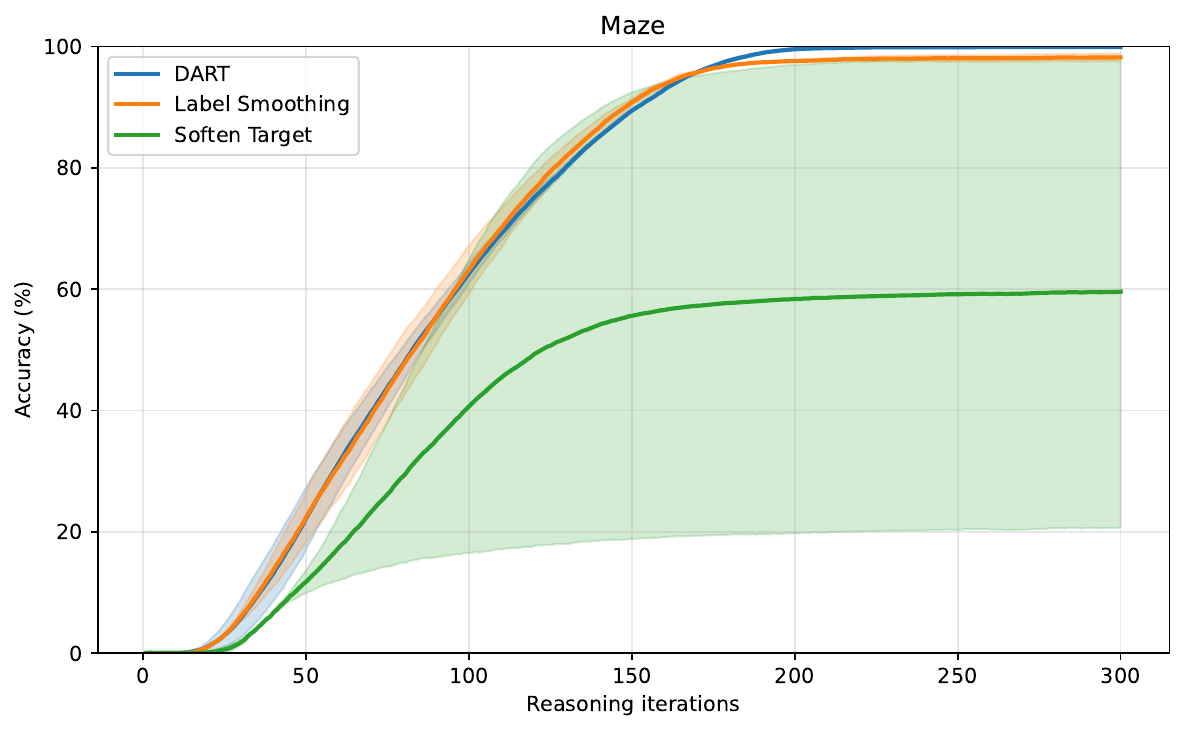}
        \caption{Target-smoothing baselines on Maze.}
        \label{fig:maze_target_smoothing}
    \end{subfigure}
    \hfill
    \begin{subfigure}{0.45\textwidth}
        \centering
        \includegraphics[width=\textwidth]{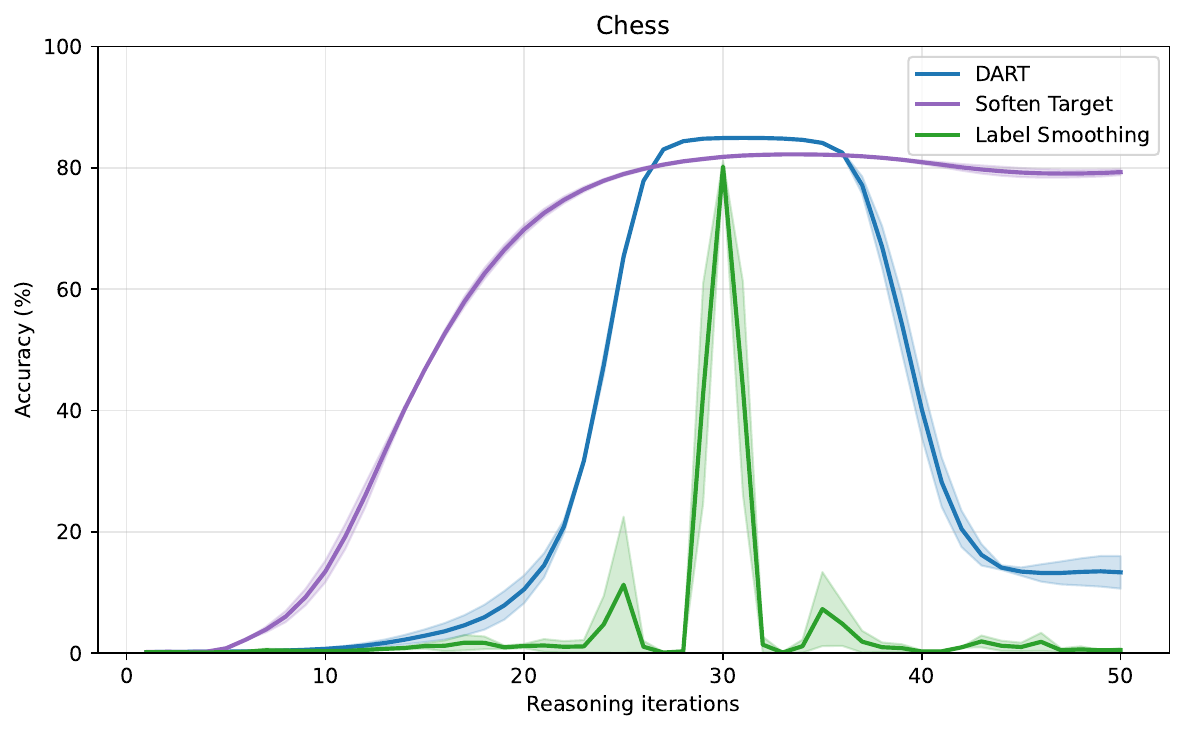}
        \caption{Target-smoothing baselines on Chess.}
        \label{fig:chess_target_smoothing}
    \end{subfigure}
    \caption{Accuracy versus reasoning iterations for target-smoothing baselines. Gaussian softened targets use the same local target construction as DART but remove the adversarial critic.}
    \label{fig:target_smoothing}
\end{figure}

\begin{table}[t]
\centering
\small
\caption{Target-smoothing baselines on Maze $31\times31$ (mean $\pm$ std over seeds).}
\label{tab:target_smoothing_maze}
\resizebox{\linewidth}{!}{%
\begin{tabular}{lccc}
\toprule
Method & Peak & Acc. @ iter 250 & Final (300) \\
\midrule
CE & $73.92 \pm 26.01$ & $73.71 \pm 26.17$ & $73.88 \pm 25.98$ \\
Label smoothing ($\epsilon=0.3$) & $98.22 \pm 0.64$ & $98.06 \pm 0.57$ & $98.21 \pm 0.65$ \\
Gaussian softened target, no critic & $59.58 \pm 38.86$ & $59.14 \pm 38.80$ & $59.58 \pm 38.86$ \\
DART & $\mathbf{99.91 \pm 0.13}$ & $\mathbf{99.89 \pm 0.15}$ & $\mathbf{99.91 \pm 0.13}$ \\
\bottomrule
\end{tabular}
}
\end{table}

\begin{table}[t]
\centering
\small
\caption{Target-smoothing baselines on Chess (mean $\pm$ std over seeds).}
\label{tab:target_smoothing_chess}
\resizebox{\linewidth}{!}{%
\begin{tabular}{lcccc}
\toprule
Method & Peak & Peak iter & Acc. @ iter 40 & Final (50) \\
\midrule
CE & $80.81 \pm 0.37$ & 30, 30 & $0.56 \pm 0.11$ & $0.92 \pm 0.08$ \\
Label smoothing ($\epsilon=0.3$) & $79.29 \pm 0.04$ & 30, 30 & $0.07 \pm 0.00$ & $0.14 \pm 0.00$ \\
Gaussian softened target, no critic & $82.22 \pm 0.09$ & 33, 34 & $\mathbf{80.93 \pm 0.28}$ & $\mathbf{79.29 \pm 0.49}$ \\
DART & $\mathbf{84.83 \pm 0.03}$ & 31, 31 & $28.58 \pm 10.47$ & $11.13 \pm 4.64$ \\
\bottomrule
\end{tabular}
}
\end{table}

\subsubsection{DART versus simple target augmentations}
Label smoothing and Gaussian softened targets isolate whether DART's gains can be explained by simpler target augmentation. Label smoothing mixes the one-hot target with a uniform distribution, reducing overconfidence but ignoring the input-specific structure of valid solutions. The Gaussian softened-target baseline uses the same local target construction as DART, but removes the adversarial critic.

On Maze $31\times31$, label smoothing helps substantially, reaching $98.22 \pm 0.64\%$ peak accuracy (Table~\ref{tab:target_smoothing_maze}). This confirms that softening brittle one-hot supervision can help CE under spatial extrapolation. However, it remains below DART, which reaches $99.91 \pm 0.13\%$ with much tighter trajectories. More importantly, Gaussian softened targets without the critic are highly unstable, reaching only $59.58 \pm 38.86\%$. Thus, simply constructing a local target neighborhood is not sufficient on Maze; the adversarial critic is needed to make that neighborhood act as a reliable, input-conditioned constraint on generated paths.

On Chess, the pattern is complementary. Label smoothing performs poorly, peaking at $79.29 \pm 0.04\%$ and collapsing almost completely by iteration 50 (Table~\ref{tab:target_smoothing_chess}). Uniformly spreading probability mass over incorrect move classes does not capture the structured ambiguity of chess decisions. Gaussian softened targets work better: they reach $82.22 \pm 0.09\%$ peak accuracy and preserve $79.29 \pm 0.49\%$ accuracy at iteration 50. This shows that local target softening can regularize recurrent dynamics and reduce collapse. Nevertheless, DART reaches a higher peak of $84.83 \pm 0.03\%$, indicating that adversarial alignment provides an additional signal beyond target smoothing alone.

Overall, the simple augmentations are useful but task-dependent. Label smoothing helps Maze but fails on Chess; Gaussian softened targets stabilize Chess but are unreliable on Maze. DART is more consistent because it combines local target relaxation with a learned input-conditioned critic, giving the model both a smoother target neighborhood and a constraint on whether its generated output distribution remains compatible with the specific problem instance.

%% file: section/limitation.tex
\section{Discussion}
\label{sec:discussion}
While RRMs with DART are highly effective on spatial algorithmic tasks such as Maze, their gains are more nuanced on games and constraint-satisfaction tasks such as Chess and Sudoku. Unlike Maze, whose difficulty scales mainly with input size and path length, these tasks are dominated by strategic depth, ambiguity, and combinatorial structure. Chess requires selecting an optimal move among many plausible candidates, while masked Sudoku involves an expanding space of possible completions that demands hypothesis generation, verification, and correction.
This exposes a key limitation of current recurrent reasoning models. Although they can iteratively refine predictions, they lack explicit mechanisms for verification, structured exploration, and backtracking. DART improves extrapolation by expanding the local soft neighborhood around the observed target and stabilizing reasoning trajectories, but it primarily mitigates instability rather than solving the underlying search problem. It also does not enumerate multiple semantically distinct valid discrete solutions; the perturbation preserves the observed target under $\arg\max$ and provides a smoother continuous supervision signal around that target. As a result, DART makes reasoning more reliable without fully eliminating the root causes of failure in complex tasks.
Our progressive-training experiments further show that long-horizon stability and easy-to-hard extrapolation are related but distinct. Progressive training is especially effective at removing late-iteration collapse on Chess, while DART is more important under the larger Maze difficulty shift. Combining distributional training with explicit curriculum, verification, or search mechanisms is therefore a natural next step.
These findings suggest that improving training objectives alone may not be sufficient for problems that inherently require search and verification. A promising direction is to combine distributional training with stronger reasoning primitives, such as learned verification modules, scoring functions, or structured backtracking. More broadly, integrating distribution-guided learning with search-based reasoning may provide a more complete framework for tackling complex reasoning tasks.

%% file: section/conclusion.tex
\section{Conclusion}

We introduced \textbf{Distributional Adversarial Recurrent Training (DART)}, a training framework for recurrent reasoning models that replaces purely instance-level supervision with distribution-guided adversarial alignment around the ground-truth solution. 
By modeling logical reasoning as a process of progressively moving predictions toward a local soft target neighborhood, DART provides a richer learning signal than standard cross-entropy and improves the stability of iterative reasoning.
Across Maze, Chess, and masked Sudoku, DART improves peak solution quality and robustness under distribution shift, with especially large gains when point-wise supervision becomes unstable. 
Additional comparisons with label smoothing, Gaussian softened targets, and progressive training show that target softening alone is task-dependent, while adversarial alignment and progressive training address complementary failure modes. These benefits appear consistently across the evaluated DTS and TRM settings, suggesting that distributional training is a promising principle for improving recurrent reasoning systems.

%% file: section/supplementary.tex
\section{Appendix}
\subsection{Datasets}
\label{appendix:datasets}

In this section, we provide detailed descriptions of the datasets used in our experiments, including their construction procedures and key properties. The Maze and Chess datasets are publicly available through the \texttt{easy-to-hard-data} Python package, while the Masked Sudoku benchmark is constructed from the Extreme Sudoku dataset \cite{wang2025hierarchical}.

\subsubsection{Maze Dataset}
Maze instances are generated using a depth-first search (DFS) algorithm. The grid is initially initialized with walls at every cell boundary. Starting from a designated cell, DFS is applied to visit all cells while removing walls along the traversal path, ensuring connectivity.

The resulting mazes exhibit non-uniform path length distributions. Although the generation process may introduce duplicate samples, the duplication rate is below 0.5\% and is therefore ignored in our analysis. Each maze is paired with a target binary mask indicating the optimal path between the start and end points.

\subsubsection{Chess Puzzle Dataset}
Chess puzzles are derived from the Lichess database and represented in Forsyth--Edwards Notation (FEN). Each board state is converted into a tensor of size $8 \times 8 \times 12$, where each channel corresponds to a specific piece type and color.

The target output is defined as the optimal move in UCI format, encoded as an $8 \times 8$ binary mask marking the origin and destination squares of the move.

The dataset is constructed by analyzing approximately 200 million games using the Stockfish 12/13 NNUE engine. While user identities are not explicitly included, they may be indirectly inferred through puzzle replay on the Lichess platform. The dataset is released under the Creative Commons CC0 license, allowing unrestricted use for research purposes.

\subsubsection{Masked Sudoku Dataset}
The masked Sudoku dataset is constructed from complete Sudoku solutions provided in the Hard Sudoku benchmark \cite{wang2025hierarchical}. Each example is represented as a $9 \times 9$ grid, where entries take values in $\{1,\dots,9\}$. To create a partially observed puzzle, we randomly mask a subset of cells and replace them with a special unknown token, while keeping the full solved grid as the target output.

We use this dataset to define a masking-based easy-to-hard generalization benchmark. During training, 1{,}000 solved grids is converted into a puzzle by masking 20\% of its cells. At evaluation time, we increase the masking ratio to 30\%, 40\%, and 50\%, thereby controlling task difficulty through the amount of missing information. Higher masking ratios make the reasoning problem harder because fewer constraints are directly observed in the input, requiring the model to infer a larger portion of the solution from global consistency.

This construction preserves the same input-output format across difficulty levels while inducing substantial distribution shift in the amount of observable information. In our experiments, we evaluate on a held-out test set of 10{,}000 masked puzzles, which allows us to assess whether recurrent reasoning models trained on easier partially observed instances can generalize to harder ones under more severe masking.

\subsection{Hyperparameters}
\label{subsec:hyper}
Table~\ref{tab:hyper} summarizes the main training hyperparameters used in our experiments. In each training run, we hold out 20\% of the training data as a validation set and select for final testing the checkpoint that achieves the highest validation accuracy. For Sudoku, we follow a slightly different model-selection protocol: because the task difficulty is controlled by the masking ratio, we evaluate checkpoints on a validation split constructed with the same masking ratio as the target test setting.

\begin{table}[h]
\centering
\caption{Training hyperparameters. Dashes indicate that we did not utilize those options.}
\begin{tabular}{lcccccccc}
\toprule
task & optim. &  lr & lr critic & decay schedule & decay factor & warm-up & epochs & clip \\
\midrule
Mazes & adamw & 1e-3 & 5e-5 & -- & -- & 10 & 50 & 0.01 \\
Chess & sgd & 1e-2 & 5e-5 & [100, 110] & 0.01 & 3 & 120 & 0.01 \\
Sudoku & adam & 1e-3 & 5e-5 & [1000] & 0.01 & 10 & 3000 & 0.01 \\
\bottomrule
\label{tab:hyper}
\end{tabular}
\end{table}

\subsection{Model architectures}
\label{appendix:model_arch}
For all DTS models on the easy-to-hard tasks, we use the same backbone architecture. The projection layer $p$ is implemented as a convolutional layer that maps the raw input to a hidden representation of dimension $d = 128$. The recurrent block $r$ is implemented as a residual block with two convolutional layers, and the prediction head $h$ is a final convolutional layer that maps the hidden state to the output space. For all tasks, we set the number of recurrent reasoning iterations during training to $T = 30$.

For TRM, we follow the architecture in the original paper, but reduce the hidden size from $512$ to $128$. During training, we use $H = 3$ high-level cycles and $L = 6$ low-level cycles. 

For DART, we set the Gaussian noise standard deviation to $\sigma = 0.3$ for all tasks. The critic is constructed to mirror the generator architecture. For DTS-based models, the critic uses the same architecture as the generator, but with the number of recurrent reasoning iterations set to $T = 5$ during training. For TRM, we use a critic with $H = 2$ high-level cycles and $L = 3$ low-level cycles.
The Wasserstein critic is constrained using weight clipping. We use a clipping threshold of $0.01$ for Maze, Chess, and Sudoku. For optimization stability, we additionally use gradient clipping with threshold $1$ for Sudoku; Maze and Chess do not use gradient clipping. The generator and critic are updated once per minibatch in alternation, and the critic learning rate is reported in Table~\ref{tab:hyper}.

We analyze the effects of different values of $\sigma$ and different critic architectures in the Ablation section \ref{sec:ablation}.

\subsection{Computational resources}
\label{appendix:comp_res}
All experiments were conducted using a NVIDIA RTX 5090 GPU. For the DTS experiments, Maze models require up to two hours. DTS models on Chess are the most computationally demanding and require between 6 and 8 hours to complete. For TRM on Sudoku, training also takes between 6 and 8 hours.

Label smoothing and Gaussian softened-target baselines modify only the target tensor and do not instantiate an additional network, so their training cost is the same as CE up to negligible target-construction overhead. DART adds the critic forward/backward pass during training. The critic mirrors the generator backbone, so trainable parameters approximately double during training, although the critic is discarded at inference time. On our DTS setup, the generator has 783{,}504 parameters on Maze and 804{,}240 parameters on Chess.

\begin{table}[h]
\centering
\small
\caption{Representative Chess training cost on one RTX 5090. GPU-hours assume 120 epochs.}
\label{tab:compute_chess}
\begin{tabular}{lccc}
\toprule
Method & Seconds / epoch & Relative to CE & GPU-hours \\
\midrule
CE & 59.0 & $1.00\times$ & 1.97 \\
Label smoothing & 59.0 & $1.00\times$ & 1.97 \\
Gaussian softened target & 59.0 & $1.00\times$ & 1.97 \\
DART & 104.4 & $1.77\times$ & 3.48 \\
Progressive CE & 83.8 & $1.42\times$ & 2.79 \\
Progressive DART & 127.4 & $2.16\times$ & 4.25 \\
\bottomrule
\end{tabular}
\end{table}

Inference cost is unchanged by DART because the critic is not used at test time. In our wall-clock measurements, Maze evaluation at 300 iterations takes 334 seconds for both CE and DART, and Chess evaluation at 50 iterations with matched batch size takes 9.5 seconds for CE and 10.0 seconds for DART.

%% file: section/ablation.tex
\section{Additional ablations}
\label{sec:ablation}
\subsection{Effect of Target Distribution Width.}

We study the effect of target distribution width in DART by varying the standard deviation $\sigma$ that defines the neighborhood around the ground-truth solution. Intuitively, $\sigma$ controls how much deviation from the exact solution is tolerated during training.
Figure~\ref{fig:mazes_sigma} shows that changing $\sigma$ from 0.1 to 0.3 has minimal effect on Maze. The reasoning trajectories remain stable across iterations for both settings, and the final accuracies are nearly identical ($99.88 \pm 0.12\%$ vs. $99.91 \pm 0.13\%$). These results indicate that performance on Maze is largely insensitive to the precise choice of $\sigma$.
The effect is more pronounced on Chess (Figure~\ref{fig:chess_sigma}). Both DART variants outperform the CE baseline by roughly 4\% in peak accuracy ($84.82 \pm 0.18\%$ and $84.93 \pm 0.03\%$ versus $80.8 \pm 0.37\%$). Moreover, performance remains consistent across $\sigma = 0.1$ and $\sigma = 0.45$, indicating that DART is effective across a broad range of target widths. However, the larger $\sigma$ produces a wider high-accuracy plateau, whereas the smaller $\sigma$ peaks more sharply and degrades earlier. This behavior suggests that broader target distributions regularize a wider region of the valid solution manifold, allowing the model to learn nearby valid intermediate states rather than only the exact target trajectory. Consequently, the iterative reasoning process becomes more robust to small accumulated errors, improving stability over long reasoning horizons and generalization to more difficult tasks.
Overall, DART remains effective across a broad range of $\sigma$ values, while larger target widths can provide additional robustness on harder reasoning problems.

\subsection{Effects of Critic Architecture}

\begin{figure}[h]
    \centering
    \begin{subfigure}{0.45\textwidth}
        \centering
        \includegraphics[width=\textwidth]{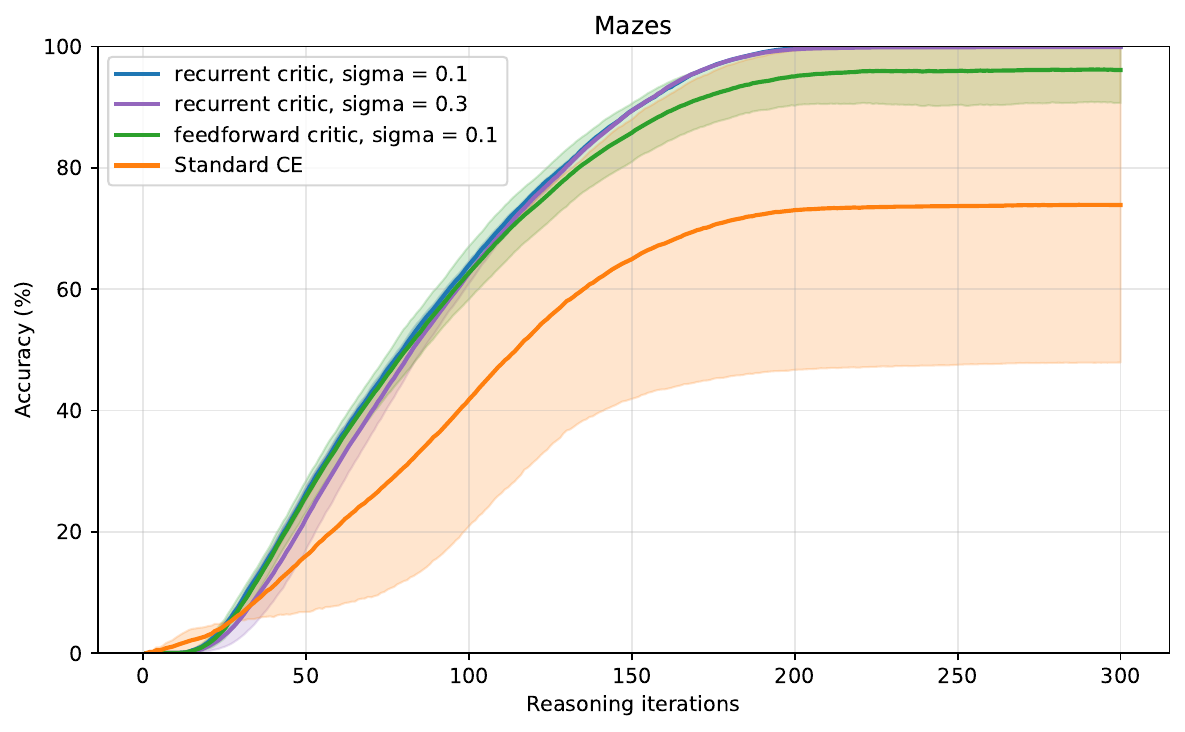}
        \caption{Maze
            }
        \label{fig:mazes_sigma}
    \end{subfigure}
    \hfill
    \begin{subfigure}{0.45\textwidth}
        \centering
        \includegraphics[width=\textwidth]{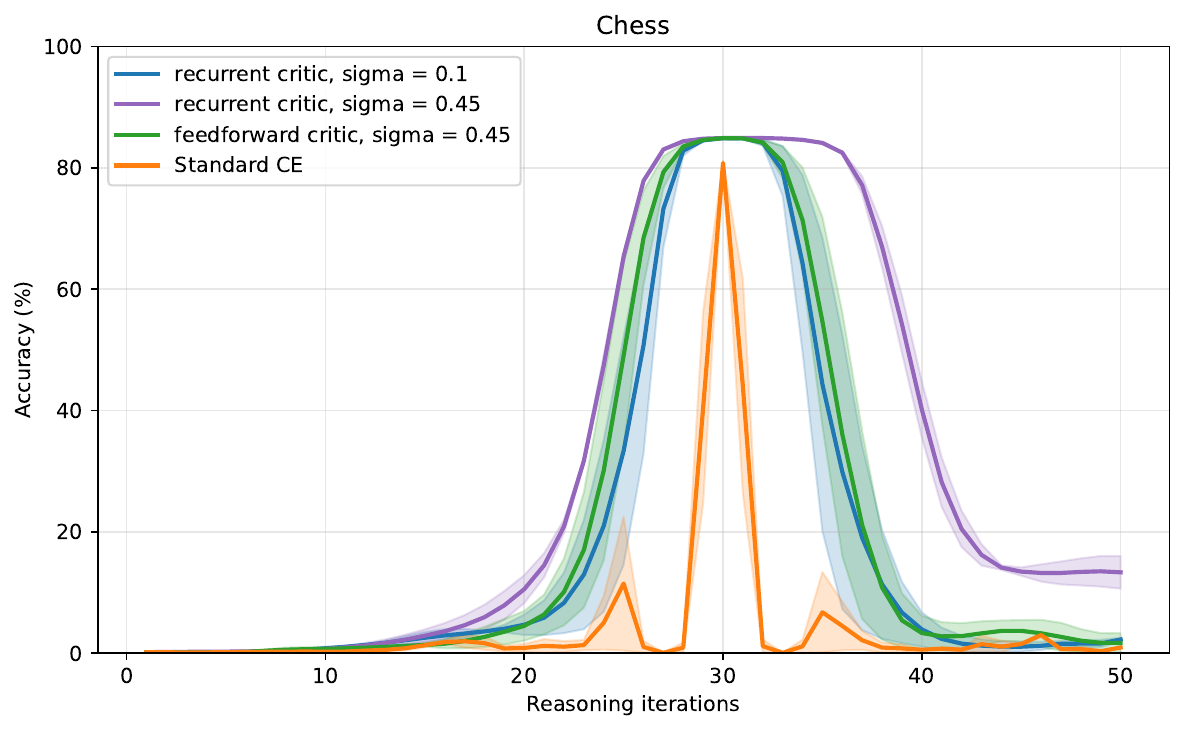}
        \caption{Chess
            }
        \label{fig:chess_sigma}
    \end{subfigure}
    \caption{Effect of the target distribution width $\sigma$ and the critic architecture on reasoning performance.}
    \label{fig:sigma_and_critic}
\end{figure}

We study the effect of critic architecture in DART by comparing a recurrent critic with a feedforward critic under the same $\sigma$. The feedforward critic uses the same residual building block as the recurrent critic, but replaces recurrent reuse with depth by stacking five independent residual blocks instead of applying one shared block for five iterations.
Figure~\ref{fig:mazes_sigma} shows that, on Maze, DART with a feedforward critic still substantially outperforms the CE baseline ($96.23 \pm 5.32\%$ vs. $73.92 \pm 26.01\%$), but remains less stable than DART with a recurrent critic ($99.88 \pm 0.12\%$) under the same setting of $\sigma = 0.1$. 
This suggests that DART does not rely on a recurrent critic to improve performance, although recurrence provides stronger regularization for the iterative reasoning process.
This pattern is even clearer on Chess (Fig.~\ref{fig:chess_sigma}). Both critic variants outperform the CE baseline, with peak accuracies of $84.90 \pm 0.03\%$ for the feedforward critic and $84.93 \pm 0.03\%$ for the recurrent critic, compared with $80.8 \pm 0.37\%$ for CE. This confirms that distributional alignment is robust across critic designs. However, the recurrent critic maintains high accuracy over a wider range of reasoning iterations, whereas the feedforward critic degrades earlier. This suggests that although a feedforward critic is sufficient to improve solution quality, the recurrent critic is better suited to stabilizing intermediate reasoning states over long computation horizons.
Overall, DART is robust to the choice of critic architecture, but recurrent critics yield more stable reasoning trajectories, especially on harder tasks. 

\subsection{Additional Sample Trajectory Analysis}

We further inspect individual reasoning trajectories to test the mechanism suggested by our hypothesis. If the adversarial critic is acting as an input-conditioned distributional constraint, then the generated trajectory should not merely hit the correct solution at one iteration; it should enter the high-density region of successful traces and remain close to it. We therefore plot both IoU and exact match: IoU captures partial overlap with the target, while exact match verifies that the discrete solution is fully correct. Conversely, a model trained only with point-wise CE may reach a high-overlap or even exact state transiently, but can drift into invalid regions because later recurrent updates are not directly constrained by the local solution neighborhood.

Figure~\ref{fig:sample_trajectory_metrics} shows this pattern on representative Maze and Chess examples. On Maze, both CE and DART first reach exact match at iteration 97, but their post-solution behavior differs. CE drops out of exact match 11 times after first solving the instance, while DART remains exact for all subsequent iterations. This supports the view that the critic does not simply accelerate convergence; it stabilizes the trajectory once it reaches the successful region.

The same effect is stronger on Chess, where overthinking is the dominant failure mode. CE reaches exact match at iteration 29 but does not retain it: only $13.6\%$ of the subsequent iterations remain exact, and the final prediction has zero IoU with the target. DART reaches exact match at iteration 27 and stays exact through the final iteration. Thus, even on a task where CE can briefly identify the correct move, the adversarial signal helps prevent the recurrent state from drifting away from the solution manifold.

The aggregate trajectory statistics in Table~\ref{tab:trajectory_stats} show that this is not an isolated example. On Maze, DART solves every evaluated sample at least once and also finishes every sample in an exact state. CE solves only 51.6\% of samples at least once and finishes at 43.8\%, indicating that many trajectories either never enter the successful region or enter it only transiently. The exact-match AUC makes this difference clearer: DART spends 68.9\% of its evaluated trajectory in an exact state on average, compared with only 28.4\% for CE.

On Chess, CE's failure mode is almost entirely a retention failure. It reaches an exact solution at least once for 84.8\% of samples, but the final exact-match rate is 0.0\%. In other words, CE often discovers a correct move during recurrent refinement, but subsequent iterations overwrite it. DART improves both discovery and retention: it solves 91.0\% of samples at least once, reaches the first exact solution earlier on average (23.1 vs. 28.0 iterations), and increases post-solution retention from 11.6\% to 63.5\%. The exact-match AUC similarly rises from 4.5\% to 32.3\%, showing that DART spends a much larger fraction of the reasoning horizon in a valid discrete state.

These qualitative trajectories are consistent with the aggregate results in Section~\ref{sec:experiment}. DART improves most when point-wise supervision becomes brittle: under large spatial extrapolation in Maze, and under late-iteration overthinking in Chess. The critic encourages generated outputs to remain in regions that resemble softened successful traces for the specific input, which provides a stabilizing force that simple one-hot supervision lacks.

The exact-match trace is important because high overlap alone can hide invalid intermediate states. In Maze, a path prediction may share most cells with the target while containing a small break or spurious branch that makes the decoded path invalid. In Chess, a prediction may assign probability mass to a plausible move neighborhood while still selecting the wrong square after arg max decoding. The exact-match metric therefore exposes whether the recurrent process has truly entered the valid discrete solution set, not merely moved close to it in overlap space.

This distinction clarifies the role of the critic. CE supplies supervision only at the selected target configuration, so once the model reaches a nearly correct state, later recurrent updates can still move it along unconstrained directions that preserve some local overlap but destroy global validity. DART instead compares generated outputs against samples from a local target neighborhood conditioned on the input. This encourages the recurrent trajectory to remain near the set of successful traces rather than repeatedly crossing the boundary between valid and invalid states. The observed post-solution stability of DART in both Maze and Chess is therefore consistent with the distributional-alignment interpretation of the method.

%% file: neurips.bib
@inproceedings{goodfellow2014gan,
  title={Generative Adversarial Nets},
  author={Goodfellow, Ian and others},
  booktitle={NeurIPS},
  year={2014}
}

@inproceedings{arjovsky2017wgan,
  title={Wasserstein Generative Adversarial Networks},
  author={Arjovsky, Martin and Chintala, Soumith and Bottou, Léon},
  booktitle={ICML},
  year={2017}
}

@misc{bachmann2025pitfallsnexttokenprediction,
      title={The pitfalls of next-token prediction}, 
      author={Gregor Bachmann and Vaishnavh Nagarajan},
      year={2025},
      eprint={2403.06963},
      archivePrefix={arXiv},
      primaryClass={cs.CL},
      url={https://arxiv.org/abs/2403.06963}, 
}

@article{brown2024large,
  title={Large language monkeys: Scaling inference compute with repeated sampling},
  author={Brown, Bradley and Juravsky, Jordan and Ehrlich, Ryan and Clark, Ronald and Le, Quoc V and R{\'e}, Christopher and Mirhoseini, Azalia},
  journal={arXiv preprint arXiv:2407.21787},
  year={2024}
}

@article{bansal2022end0to0end,
  title   = {End-to-end Algorithm Synthesis with Recurrent Networks: Logical Extrapolation Without Overthinking},
  author  = {Arpit Bansal and Avi Schwarzschild and Eitan Borgnia and Zeyad Emam and Furong Huang and Micah Goldblum and Tom Goldstein},
  year    = {2022},
  journal = {arXiv preprint arXiv: 2202.05826}
}

@inproceedings{bear2024rethinking,
  title     = {Rethinking Deep Thinking: Stable Learning of Algorithms Using Lipschitz Constraints},
  author    = {Jay Bear and Adam Prugel-Bennett and Jonathon Hare},
  booktitle = {Advances in Neural Information Processing Systems},
  volume    = {37},
  pages     = {97027--97052},
  year      = {2024}
}

@inproceedings{NEURIPS2021_3501672e,
  author    = {Schwarzschild, Avi and Borgnia, Eitan and Gupta, Arjun and Huang, Furong and Vishkin, Uzi and Goldblum, Micah and Goldstein, Tom},
  booktitle = {Advances in Neural Information Processing Systems},
  editor    = {M. Ranzato and A. Beygelzimer and Y. Dauphin and P.S. Liang and J. Wortman Vaughan},
  pages     = {6695-6706},
  publisher = {Curran Associates, Inc.},
  title     = {Can You Learn an Algorithm? Generalizing from Easy to Hard Problems with Recurrent Networks},
  url       = {https://proceedings.neurips.cc/paper_files/paper/2021/file/3501672ebc68a5524629080e3ef60aef-Paper.pdf},
  volume    = {34},
  year      = {2021}
}

@inproceedings{DBLP:journals/corr/KaiserS15,
  author    = {Lukasz Kaiser and Ilya Sutskever},
  editor    = {Yoshua Bengio and Yann LeCun},
  title     = {Neural GPUs Learn Algorithms},
  booktitle = {4th International Conference on Learning Representations, {ICLR} 2016, San Juan, Puerto Rico, May 2-4, 2016, Conference Track Proceedings},
  year      = {2016},
  url       = {http://arxiv.org/abs/1511.08228},
  bibsource = {dblp computer science bibliography, https://dblp.org}
}

@article{jolicoeur-martineau2025less,
  title   = {Less is More: Recursive Reasoning with Tiny Networks},
  author  = {Alexia Jolicoeur-Martineau},
  year    = {2025},
  journal = {arXiv preprint arXiv: 2510.04871}
}

@article{wang2025hierarchical,
  title   = {Hierarchical Reasoning Model},
  author  = {Guan Wang and Jin Li and Yuhao Sun and Xing Chen and Changling Liu and Yue Wu and Meng Lu and Sen Song and Yasin Abbasi Yadkori},
  year    = {2025},
  journal = {arXiv preprint arXiv: 2506.21734}
}

@article{wei2022chain0of0thought,
  title   = {Chain-of-Thought Prompting Elicits Reasoning in Large Language Models},
  author  = {Jason Wei and Xuezhi Wang and Dale Schuurmans and Maarten Bosma and Brian Ichter and Fei Xia and Ed Chi and Quoc Le and Denny Zhou},
  year    = {2022},
  journal = {arXiv preprint arXiv: 2201.11903}
}

@article{yao2023tree,
  title   = {Tree of Thoughts: Deliberate Problem Solving with Large Language Models},
  author  = {Shunyu Yao and Dian Yu and Jeffrey Zhao and Izhak Shafran and Thomas L. Griffiths and Yuan Cao and Karthik Narasimhan},
  year    = {2023},
  journal = {arXiv preprint arXiv: 2305.10601}
}

@article{shao2025deepseekmath0v20,
  title   = {DeepSeekMath-V2: Towards Self-Verifiable Mathematical Reasoning},
  author  = {Zhihong Shao and Yuxiang Luo and Chengda Lu and Z. Z. Ren and Jiewen Hu and Tian Ye and Zhibin Gou and Shirong Ma and Xiaokang Zhang},
  year    = {2025},
  journal = {arXiv preprint arXiv: 2511.22570}
}

@article{chollet2019measure,
  title   = {On the Measure of Intelligence},
  author  = {François Chollet},
  year    = {2019},
  journal = {arXiv preprint arXiv: 1911.01547}
}

@article{chollet2025arc0agi020,
  title   = {ARC-AGI-2: A New Challenge for Frontier AI Reasoning Systems},
  author  = {Francois Chollet and Mike Knoop and Gregory Kamradt and Bryan Landers and Henry Pinkard},
  year    = {2025},
  journal = {arXiv preprint arXiv: 2505.11831}
}

@article{graves2014neural,
  title   = {Neural Turing Machines},
  author  = {Alex Graves and Greg Wayne and Ivo Danihelka},
  year    = {2014},
  journal = {arXiv preprint arXiv: 1410.5401}
}

@misc{kingma2022autoencodingvariationalbayes,
      title={Auto-Encoding Variational Bayes}, 
      author={Diederik P Kingma and Max Welling},
      year={2022},
      eprint={1312.6114},
      archivePrefix={arXiv},
      primaryClass={stat.ML},
      url={https://arxiv.org/abs/1312.6114}, 
}

@article{sun2024easy0to0hard,
  title     = {Easy-to-Hard Generalization: Scalable Alignment Beyond Human Supervision},
  author    = {Zhiqing Sun and Longhui Yu and Yikang Shen and Weiyang Liu and Yiming Yang and S. Welleck and Chuang Gan},
  journal   = {Neural Information Processing Systems},
  year      = {2024},
  doi       = {10.48550/arXiv.2403.09472},
  bibSource = {Semantic Scholar https://www.semanticscholar.org/paper/9ccb5de1e22238b93a6af01c1dc341dc9bc3f28d}
}

@article{shu2020encoding,
  title     = {Encoding Robustness to Image Style via Adversarial Feature Perturbations},
  author    = {Manli Shu and Zuxuan Wu and Micah Goldblum and T. Goldstein},
  journal   = {Neural Information Processing Systems},
  year      = {2020},
  bibSource = {Semantic Scholar https://www.semanticscholar.org/paper/62edba38f15ab8d36626d4233e5b42f7392c392a}
}

@inproceedings{DBLP:conf/nips/VeerabadranRTRS23,
  author    = {Vijay Veerabadran and Srinivas Ravishankar and Yuan Tang and Ritik Raina and Virginia de Sa},
  editor    = {Alice Oh and Tristan Naumann and Amir Globerson and Kate Saenko and Moritz Hardt and Sergey Levine},
  title     = {Adaptive recurrent vision performs zero-shot computation scaling to unseen difficulty levels},
  booktitle = {Advances in Neural Information Processing Systems 36: Annual Conference on Neural Information Processing Systems 2023, NeurIPS 2023, New Orleans, LA, USA, December 10 - 16, 2023},
  year      = {2023},
  url       = {http://papers.nips.cc/paper\_files/paper/2023/hash/3a40e042c66e84659249f3254460c123-Abstract-Conference.html},
  bibsource = {dblp computer science bibliography, https://dblp.org}
}

@article{graves2016adaptive,
  title   = {Adaptive Computation Time for Recurrent Neural Networks},
  author  = {Alex Graves},
  year    = {2016},
  journal = {arXiv preprint arXiv: 1603.08983}
}

@article{article,
author = {Gers, Felix and Schmidhuber, E.},
year = {2001},
month = {12},
pages = {1333 - 1340},
title = {LSTM recurrent networks learn simple context-free and context-sensitive languages},
volume = {12},
journal = {Neural Networks, IEEE Transactions on},
doi = {10.1109/72.963769}
}

@article{banino2021pondernet0,
  title   = {PonderNet: Learning to Ponder},
  author  = {Andrea Banino and Jan Balaguer and Charles Blundell},
  year    = {2021},
  journal = {arXiv preprint arXiv: 2107.05407}
}

@misc{yang2024structureguidedadversarialtrainingdiffusion,
      title={Structure-Guided Adversarial Training of Diffusion Models}, 
      author={Ling Yang and Haotian Qian and Zhilong Zhang and Jingwei Liu and Bin Cui},
      year={2024},
      eprint={2402.17563},
      archivePrefix={arXiv},
      primaryClass={cs.CV},
      url={https://arxiv.org/abs/2402.17563}, 
}

@article{kaya2018shallow0deep,
  title     = {Shallow-Deep Networks: Understanding and Mitigating Network Overthinking},
  author    = {Yigitcan Kaya and Sanghyun Hong and Tudor Dumitras},
  journal   = {International Conference on Machine Learning},
  year      = {2018},
  bibSource = {Semantic Scholar https://www.semanticscholar.org/paper/a3143eaa68040d366848a9c324b29d3f56f97a5d}
}

@inproceedings{Eyzaguirre_2020_CVPR,
  author    = {Eyzaguirre, Cristobal and Soto, Alvaro},
  title     = {Differentiable Adaptive Computation Time for Visual Reasoning},
  booktitle = {Proceedings of the IEEE/CVF Conference on Computer Vision and Pattern Recognition (CVPR)},
  month     = {June},
  year      = {2020}
}

@article{dehghani2018universal,
  title   = {Universal Transformers},
  author  = {Mostafa Dehghani and Stephan Gouws and Oriol Vinyals and Jakob Uszkoreit and Łukasz Kaiser},
  year    = {2018},
  journal = {arXiv preprint arXiv: 1807.03819}
}
